%% file: templateArxiv.tex
\documentclass{article}

\usepackage{PRIMEarxiv}

\usepackage{hyperref}
\hypersetup{hidelinks}
\usepackage{booktabs}
\usepackage{graphicx}
\usepackage{wrapfig}
\usepackage{multirow} 
\usepackage{amsmath,amssymb}
\usepackage{array,makecell,float}

\usepackage[utf8]{inputenc} 
\usepackage[T1]{fontenc}    
\usepackage{hyperref}       
\usepackage{url}            
\usepackage{booktabs}       
\usepackage{amsfonts}       
\usepackage{nicefrac}       
\usepackage{microtype}      
\usepackage{lipsum}
\usepackage{fancyhdr}       
\usepackage{graphicx}       
\graphicspath{{media/}}     

\title{SpikeLite: Lightweight Spiking Neural\\ Networks for Time-Series Forecasting
}

\author{
  Bang Hu, Changze Lv, Mingjie Li, Xiaoqing Zheng, Wei cao, Fan Zhang \\
  School of Computer Science, Fudan University, Shanghai, China\\
}

\begin{document}
\maketitle

\begin{abstract}
Spiking neural networks (SNNs) offer an energy-efficient paradigm for time-series forecasting through spike-driven computation.
However, recent SNN forecasters often pursue higher accuracy through increasingly complex attention mechanisms, or specialized neuronal dynamics, weakening the lightweight motivation of SNNs. 
We introduce SpikeLite, a spiking forecasting framework built around two modules: a Frequency-Selective Spiking Encoder (FSSE) for frequency-sensitive temporal encoding and a Sparse Spiking Channel Attention (SSCA) module for selective cross-channel interaction.
FSSE exploits the low-pass filtering behavior of LIF dynamics to reorganize each input sequence into frequency-sensitive components while collectively preserving the input at the decomposition stage.
SSCA then learns a binary mask from encoded channel representations and uses it to selectively exchange information within spike-driven self-attention, retaining informative cross-channel interactions while suppressing redundant ones. 
When explicit channel interaction is unnecessary, SpikeLite uses the lighter FSSE-only channel-independent path.
Experiments under the SeqSNN and SpikF protocols cover four standard multivariate and eight long-term forecasting benchmarks. 
SpikeLite achieves the best aggregate performance under both protocols, with an average $R^2$ of 0.790 and RSE of 0.440, and lowest average MSE/MAE of 0.343/0.345 in long-term forecasting. 
Moreover, evaluation on the ECL dataset shows that SpikeLite achieves the lowest reported energy consumption, further demonstrating its potential for energy-efficient time-series forecasting.
\end{abstract}

\section{Introduction}
Spiking neural networks (SNNs) are regarded as the third generation of neural networks, emulate biological neuronal dynamics and communicate through discrete spike events, 
offering a biologically inspired and potentially energy-efficient alternative to artificial neural networks (ANNs )~\cite{maass1997networks}.
Recent advances have extended SNNs to many tasks, including image classification~\cite{fang2021deep,zhou2023spikformer}, object detection~\cite{kim2020spikingyolo,su2023deep}, semantic segmentation~\cite{kim2022beyond}, speech recognition~\cite{wu2020deep,pellegrini2021low}, and autonomous driving and control~\cite{zhu2024autonomous}.

Multivariate time-series forecasting is an important predictive task with applications in energy systems, traffic management, weather prediction, and financial analysis~\cite{qiu2025duet,han2024softs}.
Motivated by the energy-efficiency of SNNs, recent studies have begun to develop SNN-based forecasting models.
SeqSNN was the first work to systematically apply SNNs to time-series forecasting, evaluating spiking convolutional, recurrent, and Transformer architectures within a unified forecasting framework~\cite{lv2024efficient}.
Following this direction, SpikF introduces frequency-domain selection for long-term forecasting~\cite{wu2025spikf}, TS-LIF develops a dual-compartment spiking neuron to model multi-scale temporal dynamics~\cite{feng2025tslif}, and SpikeSTAG combines graph learning with spike-based temporal processing for spatio-temporal forecasting~\cite{hu2025spikestag}.
These advances improve the forecasting accuracy of SNNs, but increasingly complex neuronal dynamics, attention modules, and spatial--temporal backbones also add architectural and computational complexity, potentially compromising the energy-efficiency advantage that motivates spike-based modeling.
This raises a central question: \textbf{how can SNN-based forecasters preserve their accuracy gains while maintaining energy-efficient computation?}

We revisit this question through two fundamental aspects of multivariate forecasting: input-channel representation and cross-channel modeling. 
Existing spiking forecasters primarily focus on converting continuous observations into spike sequences, while making limited use of the latent characteristics embedded in the observed series. 
In particular, low-frequency variations often reflect slowly evolving trends, whereas higher-frequency variations capture rapid local changes and shorter-period patterns~\cite{wu2021autoformer,zhou2022fedformer}.
At the same time, modeling relationships among variables requires more than dense interactions among all channels: useful information should be exchanged selectively while redundant channel connections are suppressed.

Based on these considerations, we introduce SpikeLite, a lightweight spiking framework for time-series forecasting that combines frequency-sensitive encoding with selective cross-channel interaction. 
Its Frequency-Selective Spiking Encoder (FSSE) uses parallel LIF branches with learnable decay factors to organize each input sequence into frequency-sensitive temporal components while collectively preserving the original signal.
The Sparse Spiking Channel Attention (SSCA) module learns a binary channel-interaction mask to retain informative spike-driven exchanges and suppress redundant connections.
We evaluate the model under the SeqSNN and SpikF protocols on four standard and eight long-term forecasting benchmarks, where SpikeLite achieves the strongest overall performance among spiking forecasters while requiring the lowest estimated energy consumption.
Our contributions are summarized as follows:
\begin{itemize}
    \item We introduce SpikeLite, a lightweight spiking forecasting framework that combines frequency-sensitive input representation with selective cross-channel interaction.

    \item We develop FSSE, which exploits heterogeneous LIF dynamics to construct frequency-sensitive components and aggregate them into compact channel representations, and SSCA, which uses a learned binary mask to suppress redundant spike-driven channel interactions.

    \item We establish comprehensive evaluations under two spiking forecasting protocols. It achieves the best aggregate accuracy among the compared ANN and SNN methods and the lowest estimated energy consumption in the reported energy comparison.
\end{itemize}

\section{Related Work}
\label{sec:related_work}

\subsection{Input-Channel Representation}

Input-channel representation determines what information from the historical window is available for forecasting. 
Conventional models process historical observations through temporal convolutions, recurrent updates, or direct linear projections~\cite{lai2018modeling,bai2018empirical,zeng2023dlinear}.
To capture more complex temporal dependencies, other methods adopt elaborate Transformer backbones~\cite{wu2021autoformer,zhou2022fedformer,nie2023patchtst}.
Although deeper backbones improve modeling capacity, recent work shows that lightweight representations can remain highly competitive.
LTSF-Linear directly maps the lookback window to the forecast horizon, while DLinear improves this design by separately projecting moving-average trends and residual components~\cite{zeng2023dlinear}.
Other methods organize the input through progressive decomposition and autocorrelation~\cite{wu2021autoformer}, Fourier or wavelet representations~\cite{zhou2022fedformer}, or local temporal patches~\cite{nie2023patchtst}. 
However, such representations are designed for conventional continuous-valued networks rather than spiking models.

Spiking forecasters additionally need to encode continuous observations into discrete spike trains for spike-driven processing. 
Existing methods have successfully developed spiking representations of time-series inputs through temporal encoding and neuronal dynamics\cite{lv2024efficient}. 
However, their primary focus is on how continuous observations are converted into effective spike sequences, while the intrinsic characteristics already present in the observed signals remain less explored. 
SpikeLite addresses this limitation through FSSE, which incorporates frequency characteristics into the spiking encoding process and produces frequency-sensitive representations for each input channel.

\subsection{Cross-Channel Modeling}

After individual channel representations are constructed, forecasting models differ in how they exchange information across variables.
Channel-independent designs, such as DLinear and PatchTST, process each variable separately, often using a shared predictor structure or shared parameters across channels~\cite{zeng2023dlinear,nie2023patchtst}.
This strategy provides a simple and efficient prediction path, but may overlook useful cross-variable dependencies.
Channel-dependent designs instead introduce explicit interactions among variables through convolution, graph propagation, or attention.
For example, Crossformer uses two-stage attention to capture cross-dimension dependencies~\cite{zhang2023crossformer}, while iTransformer treats variables as tokens and applies self-attention to learn their multivariate correlations~\cite{liu2024itransformer}.

Dense interaction, however, assumes that all channel pairs are equally useful and can propagate irrelevant information. 
Prior studies show that indiscriminate mixing may introduce noise or oversmoothing, motivating mechanisms that select informative relationships~\cite{chen2024similarity,huang2023crossgnn}. 
SNN-based forecasters also support cross-channel modeling: SeqSNN applies spiking self-attention to channel-wise embeddings~\cite{lv2024efficient}, and SpikeSTAG combines adaptive graph learning with spiking temporal processing for spatio-temporal forecasting~\cite{hu2025spikestag}. 
These methods demonstrate the value of variable interaction, but their interaction structures are either dense or tied to specialized graph architectures. 
SpikeLite instead introduces SSCA, which learns a binary mask over channel pairs and uses it to retain informative spike-driven interactions while suppressing redundant connections.

\section{Preliminaries}
\label{sec:preliminary}

\subsection{Problem Formulation}

We first define the multivariate time-series forecasting task.
Given a historical multivariate sequence $\mathbf{X}=[\mathbf{x}_{1},\ldots,\mathbf{x}_{L}]^{\top} \in\mathbb{R}^{L\times C}$, where $L$ is the lookback length, $C$ is the number of channels, and $\mathbf{x}_{l}\in\mathbb{R}^{C}$ contains all channel observations at position $l$, multivariate time-series forecasting aims to predict the next $P$ observations, $\mathbf{Y}=[\mathbf{x}_{L+1},\ldots,\mathbf{x}_{L+P}]^{\top} \in\mathbb{R}^{P\times C}$.
A forecasting model $f_{\theta}$ with learnable parameters $\theta$ maps the historical sequence to the prediction $\widehat{\mathbf{Y}}=f_{\theta}(\mathbf{X})$.
Throughout the paper, $l$ indexes the original observation sequence, whereas $t\in\{1,\ldots,T_s\}$ indexes the internal simulation steps of a spiking layer.

\subsection{Spiking Neural Computation}

Spiking neural networks (SNNs) model the evolution of neuronal membrane potentials and transmit information through discrete spike events. 
Among the various spiking neuron models, the leaky integrate-and-fire (LIF) neuron is widely adopted because it provides a simple mechanism for capturing temporal state dynamics.
Given an input current $\mathbf{I}^{t}$ at simulation step $t$, an LIF neuron performs leaky integration, thresholding, and reset according to
\begin{equation}
\left\{
\begin{aligned}
U^{t} &= \lambda U^{t-1}+\mathbf{I}^{t}
        -\vartheta \mathbf{S}^{t-1},\\
\mathbf{S}^{t} &= \Theta\!\left(U^{t}-\vartheta\right),\\
\mathbf{G}^{t} &= U^{t}\mathbf{S}^{t}.
\end{aligned}
\right.
\end{equation}
Here, $\lambda$ denotes the membrane decay factor.
The last term implements a delayed subtractive reset.
During training, the derivative of the Heaviside function is approximated with a surrogate gradient.

Based on these neuronal dynamics, spiking self-attention (SSA) extends pairwise representation interaction to spike-driven networks~\cite{zhou2023spikformer,lv2024efficient}.
Given an input representation $\mathbf{X}$, three learnable projections followed by spiking neurons produce binary query, key, and value representations,
\begin{equation}
    \mathbf{Q}=\operatorname{SN}(\mathbf{X}\mathbf{W}_{Q}),\quad
    \mathbf{K}=\operatorname{SN}(\mathbf{X}\mathbf{W}_{K}),\quad
    \mathbf{V}=\operatorname{SN}(\mathbf{X}\mathbf{W}_{V}),
\end{equation}
where $\operatorname{SN}(\cdot)$ denotes the spiking-neuron activation. SSA then aggregates pairwise interactions among these spike representations without relying on the softmax normalization used in conventional self-attention:
\begin{equation}
    \operatorname{SSA}(\mathbf{Q},\mathbf{K},\mathbf{V})
    =
    \operatorname{SN}\!\left(
    \alpha\,\mathbf{Q}\mathbf{K}^{\top}\mathbf{V}
    \right),
\end{equation}
where $\alpha$ scales the magnitude of the aggregated interactions to maintain an appropriate activation range for the subsequent spiking neuron.

\section{Method}
\label{sec:method}

\begin{figure*}[t]
    \centering
    \includegraphics[width=0.95\textwidth]{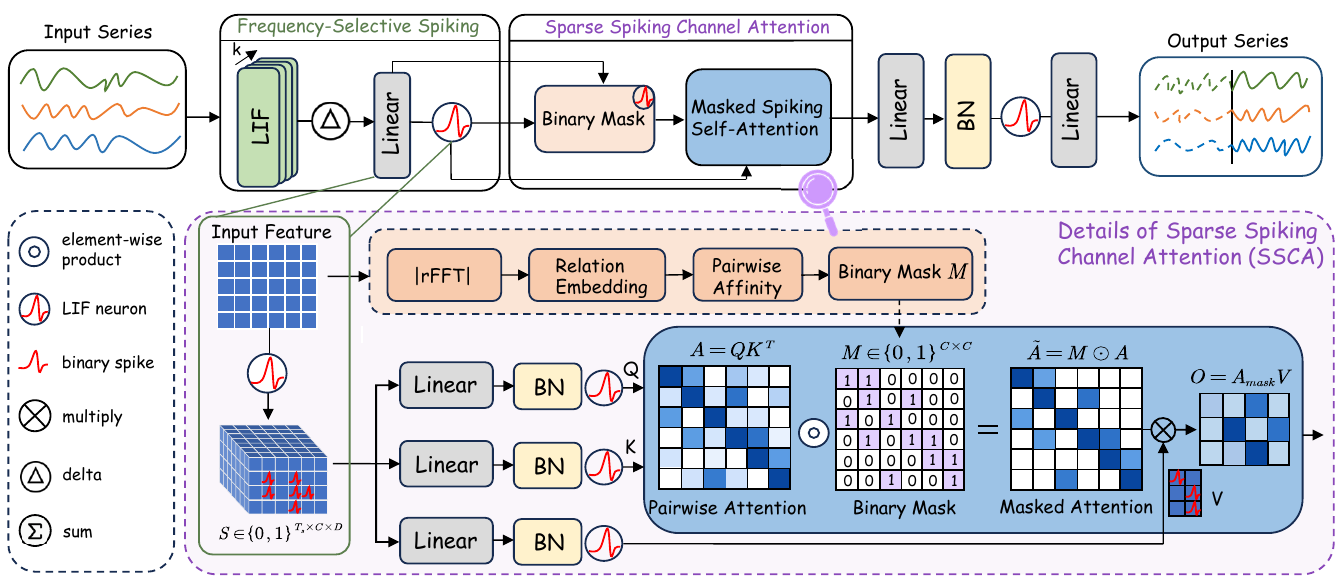}
    \caption{
    Overview of SpikeLite.
    FSSE constructs frequency-sensitive representations for individual input channels, while SSCA selectively exchanges information across channels.
    The lower panel details mask generation and masked spiking attention: the mask is estimated from the encoded channel representations and applied to the attention scores before value aggregation.
    }
    \label{fig:framework}
\end{figure*}

\subsection{Overview}

As illustrated in Fig.~\ref{fig:framework}, SpikeLite is a lightweight spiking framework for multivariate time-series forecasting that combines the Frequency-Selective Spiking Encoder (FSSE) with Sparse Spiking Channel Attention (SSCA).
Given $\mathbf{X}\in\mathbb{R}^{L\times C}$, FSSE applies parallel LIF dynamics with learnable decay factors to each input channel, and reorganizes their event-gated responses into a compact representation $\mathbf{Z}\in\mathbb{R}^{C\times D}$, where $\mathbf{Z}_{c,:}$ represents the $c$-th channel.
FSSE then repeats $\mathbf{Z}$ over $T_s$ simulation steps and converts it into binary spike states $\mathbf{S}_{\mathrm{in}}^{(t)}\in\{0,1\}^{C\times D}$.
SSCA uses a learned binary interaction mask to selectively exchange information across channels while suppressing redundant connections; when omitted, the model follows the lighter FSSE-only path.
A lightweight prediction head maps the resulting representation to $\widehat{\mathbf{Y}}=g_{\psi}(\widetilde{\mathbf{Z}}) \in\mathbb{R}^{P\times C}$.

\subsection{Frequency-Selective Spiking Encoder}
\label{sec:fsse}

\begin{wrapfigure}{r}{0.48\textwidth}
    \centering
    \vspace{-8pt}
    \includegraphics[width=\linewidth]{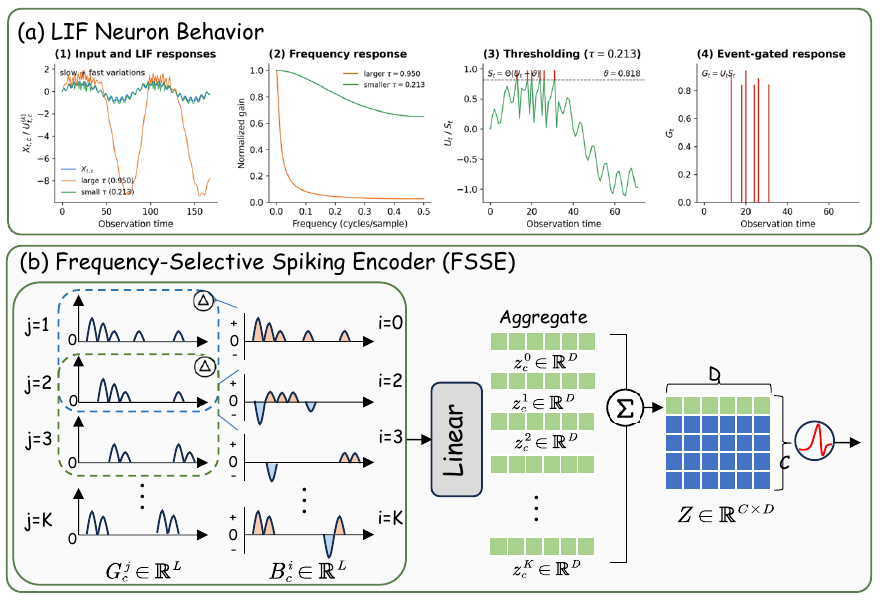}
    \caption{LIF responses and the FSSE encoding process.
    (a) LIF responses under different decay factors.
    (b) Details of Frequency-Selective Spiking Encoder.}
    \label{fig:fsse_detail}
    \vspace{-10pt}
\end{wrapfigure}

\paragraph{Frequency-selective LIF dynamics.}
As defined in Section~\ref{sec:preliminary}, the LIF neuron integrates the current input with its decayed membrane state. 
To analyze this dynamics in the frequency domain, we temporarily remove thresholding and reset, yielding the linearized equation
$U_l=\tau U_{l-1}+X_l$.
Its $z$-transform gives
$H(z)=U(z)/X(z)=1/(1-\tau z^{-1})$.
Evaluating this transfer function on the unit circle,
$z=e^{\mathrm{i}\omega}$, produces
\begin{equation}
H_{\tau}(e^{\mathrm{i}\omega})
=
\frac{1}{1-\tau e^{-\mathrm{i}\omega}},
\qquad
\left|H_{\tau}(e^{\mathrm{i}\omega})\right|
=
\frac{1}{\sqrt{1+\tau^2-2\tau\cos\omega}},
\label{eq:fsse_filter}
\end{equation}
where $\omega$ denotes the angular frequency.
Thus, leaky integration behaves as a first-order low-pass filter: a larger $\tau$ more strongly favors low-frequency components, whereas a smaller $\tau$ produces a flatter response and retains a broader frequency range.
Different decay factors therefore induce distinct frequency sensitivities to the same input, as illustrated in Fig.~\ref{fig:fsse_detail}.

The complete LIF neuron restores thresholding and reset, making the response event-dependent. 
FSSE instantiates $K$ parallel branches for each input variable, with branch-specific learnable decay factors.
For branch $k$ and variable $c$, the dynamics along the original observation axis are
  \begin{equation}
      U_{l,c}^{(k)}
      =
      \tau_{c}^{(k)}U_{l-1,c}^{(k)}
      +
      X_{l,c}
      -
      \vartheta_{c}^{(k)}S_{l-1,c}^{(k)},
      \label{eq:fsse_state}
  \end{equation}
  \begin{equation}
      S_{l,c}^{(k)}
      =
      \Theta\!\left(
          U_{l,c}^{(k)}-\vartheta_{c}^{(k)}
      \right),
      \label{eq:fsse_spike}
  \end{equation}
where $U_{l,c}^{(k)}$ is the membrane state, $S_{l,c}^{(k)}\in\{0,1\}$ is the emitted event, and $\vartheta_{c}^{(k)}$ is the firing threshold. 
Both $U_{0,c}^{(k)}$ and $S_{0,c}^{(k)}$ are initialized to zero. 
The last term in Eq.(5) implements the delayed subtractive reset.

FSSE gates each membrane response with its emitted spike:
\(
G_{l,c}^{(k)}=U_{l,c}^{(k)}S_{l,c}^{(k)}.
\)
Thus, $G_{l,c}^{(k)}$ is zero when branch $k$ is inactive and retains the corresponding membrane state when the branch fires, combining the frequency-selective neuronal dynamics with event-based transmission.

Each decay factor is parameterized as
\(
\tau_c^{(k)}=\sigma(\rho_c^{(k)})
\),
where $\rho_c^{(k)}$ is trainable and $\sigma$ denotes the logistic sigmoid. 
We initialize the branches with descending decay factors to establish a nominal slow-to-fast ordering. During training, the decay factors are freely optimized without any explicit constraint on their relative ordering. Despite the absence of an explicit ordering constraint, the learned decay factors remain broadly consistent with the initial slow-to-fast ordering.

\paragraph{Frequency-sensitive encoding.}
To obtain informative frequency-sensitive components without discarding the original signal, FSSE applies successive differences to the event-gated responses of the parallel LIF branches. 
Let $\mathbf{G}^{(k)}\in\mathbb{R}^{L\times C}$ collect the responses of branch $k$ over the historical window.
FSSE constructs $K+1$ components as
\begin{equation}
    \mathbf{B}^{(j)}
    =
    \begin{cases}
        \mathbf{G}^{(1)},
        & j=0,\\
        \mathbf{G}^{(j+1)}-\mathbf{G}^{(j)},
        & 1\leq j<K,\\
        \mathbf{X}-\mathbf{G}^{(K)},
        & j=K.
    \end{cases}
    \label{eq:fsse_components}
\end{equation}
The first component retains the response of the first LIF branch, the intermediate components describe the response differences between adjacent branches, and the final component preserves the residual between the input and the last branch response.
Because neighboring branch responses are subtracted successively, the components form a telescoping decomposition:
\begin{equation}
\begin{aligned}
    \sum_{j=0}^{K}\mathbf{B}^{(j)}
    =
    \mathbf{G}^{(1)}
    +
    \sum_{j=1}^{K-1}
    \left(
        \mathbf{G}^{(j+1)}-\mathbf{G}^{(j)}
    \right)
    +
    \left(
        \mathbf{X}-\mathbf{G}^{(K)}
    \right)
    =
    \mathbf{X}.
\end{aligned}
\label{eq:fsse_reconstruction}
\end{equation}

The decomposition in
Eq.(8) produces the components
$\{\mathbf{B}^{(j)}\}_{j=0}^{K}$ while preserving the input collectively.
Each component is independently projected along the observation axis and
aggregated into a compact channel representation:
\begin{equation}
\begin{aligned}
\mathbf{z}_{c}^{(j)}
&=
\mathbf{W}^{(j)}
\mathbf{B}_{:,c}^{(j)}
+
\mathbf{b}^{(j)}
\in\mathbb{R}^{D},
\\
\mathbf{Z}_{c,:}
&=
\sum_{j=0}^{K}
\alpha_j\mathbf{z}_{c}^{(j)},
\qquad
\mathbf{Z}\in\mathbb{R}^{C\times D},
\end{aligned}
\label{eq:fsse_projection_aggregation}
\end{equation}
where $\mathbf{W}^{(j)}\in\mathbb{R}^{D\times L}$ and $\alpha_j$ is a
learnable scale for component $j$. The resulting $\mathbf{Z}$ is passed to
the spike-driven interaction stage described in Section~\ref{sec:ssca}.

\subsection{Sparse Spiking Channel Attention}
\label{sec:ssca}

FSSE produces frequency-sensitive representations for each input variable while collectively preserving the original signal.
Because forecasting can also depend on relationships among variables, SSCA selectively exchanges information across these representations.
A standard spiking self-attention block considers all channel pairs, whereas SSCA learns a sample-dependent binary mask $\mathbf{M}\in\{0,1\}^{C\times C}$ to retain informative connections during spike-driven aggregation and suppress redundant interactions.

\paragraph{Binary mask generation.}
Before spike replication, SSCA applies a real-valued Fourier transform to
each row of $\mathbf{Z}$ along the representation dimension:
\begin{equation}
\mathbf{F}_{c}
=
\left|
\operatorname{rFFT}
\left(
\mathbf{Z}_{c,:}
\right)
\right|
\in\mathbb{R}^{F},
\qquad
F=\left\lfloor\frac{D}{2}\right\rfloor+1.
\label{eq:ssca_fft}
\end{equation}
These descriptors are used only to estimate the binary channel-interaction
mask. The subsequent attention operation is performed on the spike-form
channel states obtained after replicating $\mathbf{Z}$ over the $T_s$
simulation steps, rather than on the spectral descriptors.

To measure the similarity between channel descriptors, SSCA maps each
spectral descriptor into a low-dimensional relation space using a trainable
projection matrix:
\begin{equation}
\mathbf{r}_{c}
=
\mathbf{W}_{r}\mathbf{F}_{c}
\in\mathbb{R}^{r},
\qquad
\mathbf{W}_{r}\in\mathbb{R}^{r\times F}.
\label{eq:ssca_relation_projection}
\end{equation}
For a pair of channels $(i,j)$, we compute the squared relation distance
and its inverse affinity:
\begin{equation}
d_{ij}
=
\left\|
\mathbf{r}_{i}-\mathbf{r}_{j}
\right\|_{2}^{2}
+\epsilon,
\qquad
a_{ij}=d_{ij}^{-1}.
\label{eq:ssca_pairwise_distance}
\end{equation}
The affinity is normalized independently for each source channel:
\begin{equation}
p_{ij}
=
\operatorname{clip}
\left(
\gamma
\frac{a_{ij}}
{\max_{q\neq i}a_{iq}},
0,1
\right),
\qquad i\neq j.
\label{eq:ssca_affinity}
\end{equation}
Channels with similar spectral descriptors therefore obtain larger affinity
values and are more likely to remain connected.

We obtain the binary interaction mask by thresholding the normalized
affinity while retaining all self-connections:
\begin{equation}
\overline{m}_{ij}
=
\begin{cases}
1, & i=j,\\
\mathbb{I}(p_{ij}>\eta), & i\neq j.
\end{cases}
\label{eq:ssca_binary_mask}
\end{equation}
To optimize this discrete mask, we use a straight-through estimator:
\begin{equation}
m_{ij}
=
\begin{cases}
1, & i=j,\\
\operatorname{sg}
\left(
\overline{m}_{ij}-p_{ij}
\right)
+p_{ij}, & i\neq j,
\end{cases}
\qquad
\mathbf{M}=[m_{ij}],
\label{eq:ssca_mask_learning}
\end{equation}
where $\eta$ is the sparsification threshold and
$\operatorname{sg}(\cdot)$ denotes the stop-gradient operator. The forward
value of $\mathbf{M}$ is binary, while its backward gradient is propagated
through the continuous affinity $p_{ij}$.

\paragraph{Masked spike-driven attention.}
The spike-form variable states produced by FSSE are used as the input tokens of sparse SSA. 
For attention head $m$ and simulation step $t$, the corresponding queries, keys, and values are denoted by $\mathbf{Q}^{(t,m)}$, $\mathbf{K}^{(t,m)}$, and $\mathbf{V}^{(t,m)}$.
SSCA applies the learned mask directly to the pairwise interaction matrix:
\begin{equation}
    \mathbf{A}_{\mathrm{SSCA}}^{(t,m)}
    =
    \kappa
    \left(
        \mathbf{Q}^{(t,m)}
        \mathbf{K}^{(t,m)\top}
    \right)
    \odot \mathbf{M},
    \qquad
    \mathbf{O}^{(t,m)}
    =
    \mathbf{A}_{\mathrm{SSCA}}^{(t,m)}
    \mathbf{V}^{(t,m)}.
    \label{eq:ssca_masked_attention}
\end{equation}
where $\kappa$ is the attention scaling factor.
Consistent with spike-driven SSA, no softmax normalization is applied. 
A zero entry $M_{ij}=0$ removes the contribution from variable $j$ to variable $i$, whereas $M_{ij}=1$ preserves the corresponding interaction.

The masked outputs across all attention heads and simulation steps are aggregated through the SSA output projection, yielding the channel-interaction-enhanced representation $\widetilde{\mathbf Z}$. 
This representation is subsequently mapped by the prediction head to the final forecast $\widehat{\mathbf{Y}}\in\mathbb{R}^{P\times C}$.

\section{Experiments}
\label{sec:experiments}

\subsection{Experimental Settings}
\label{sec:experimental_settings}

We evaluate SpikeLite under two representative forecasting settings: the standard multivariate forecasting protocol introduced by SeqSNN~\cite{lv2024efficient} and the long-term forecasting protocol adopted by SpikF~\cite{wu2025spikf}.
Across the two settings, we compare SpikeLite with a broad collection of ANN- and SNN-based forecasting models on datasets covering traffic, electricity, weather, and other application domains.

\paragraph{Protocols and metrics.}
Under the SeqSNN protocol~\cite{lv2024efficient}, we evaluate standard multivariate forecasting with prediction horizons of $6$, $24$, $48$, and $96$, using $R^2$ and elative squared error (RSE) as evaluation metrics. Under the SpikF protocol~\cite{wu2025spikf}, we evaluate long-term forecasting with a lookback length of $96$ and prediction horizons of $96$, $192$, $336$, and $720$, using mean squared error (MSE) and mean absolute error (MAE). Main-text results are averaged over the four prediction horizons, while the complete horizon-wise results are provided in Appendix~\ref{app:full_ltsf_results}. Dataset statistics and complete experimental configurations are summarized in Appendix~\ref{app:experimental_details}.

\begin{table*}[t]
\centering
\caption{
Experimental results on standard multivariate forecasting under the SeqSNN protocol~\cite{lv2024efficient}.
\textbf{Bold} and \underline{underlined} values denote the best and second-best results, respectively.
$\uparrow$/$\downarrow$ indicates that higher/lower values are better.
}
\label{tab:seqsnn_full}
\scriptsize
\setlength{\tabcolsep}{1.6pt}
\renewcommand{\arraystretch}{1.04}
\resizebox{\linewidth}{!}{%
\begin{tabular}{@{}l@{\hspace{2pt}}c@{\hspace{1pt}}*{16}{c}cc@{}}
\toprule
Method & Metric & \multicolumn{4}{c}{METR-LA} & \multicolumn{4}{c}{PEMS-BAY} & \multicolumn{4}{c}{Solar} & \multicolumn{4}{c}{Electricity} & \multicolumn{2}{c}{Summary}\\
\cmidrule(lr){3-6}\cmidrule(lr){7-10}\cmidrule(lr){11-14}\cmidrule(lr){15-18}\cmidrule(l){19-20}
 & & 6&24&48&96&6&24&48&96&6&24&48&96&6&24&48&96&Avg.&Rank\\
\midrule
ARIMA & $R^2\uparrow$ & .687&.441&.282&.265&.741&.723&.692&\underline{.670}&.951&.847&.725&.689&.963&.960&.914&.863&.713&9.69\\
 & RSE$\downarrow$ & .575&.742&.889&.902&.532&.548&.562&.612&.202&.365&.588&.589&.522&.534&.564&.599&.583&9.44\\
\midrule
GP & $R^2\uparrow$ & .685&.437&.265&.233&.732&.712&.689&.665&.944&.836&.711&.675&.962&.968&.912&.852&.705&11.03\\
 & RSE$\downarrow$ & .572&.738&.912&.925&.544&.532&.577&.592&.225&.388&.612&.575&.603&.612&.633&.642&.605&10.28\\
\midrule
Autoformer & $R^2\uparrow$ & .762&.548&.411&.282&.782&.711&.689&.668&.960&.852&.791&.701&.980&.977&.975&.963&.753&8.00\\
 & RSE$\downarrow$ & .569&.692&.785&.872&.452&.543&.577&\textbf{.565}&.212&.432&.622&.685&.481&.506&.566&.548&.569&8.91\\
\midrule
PatchTST & $R^2\uparrow$ & .819&.618&.434&.284&.873&.718&.686&.612&\underline{.962}&.868&.795&.729&.980&.977&.975&.963&.768&6.50\\
 & RSE$\downarrow$ & .456&.666&.792&.875&.385&.532&.569&.577&.201&.369&.467&.588&.264&.325&.326&.441&.490&6.41\\
\midrule
AGCRN & $R^2\uparrow$ & .764&.562&.442&.287&.787&.709&.694&.659&.958&.854&.788&.704&.987&.982&.978&.966&.758&6.66\\
 & RSE$\downarrow$ & .511&.696&.785&.874&.543&.543&.566&\underline{.574}&.207&.432&.625&.656&.223&.265&.296&.531&.520&7.56\\
\midrule
STAEformer & $R^2\uparrow$ & .774&.573&.396&.279&.803&.716&.691&.647&.961&.857&.790&.713&.977&.973&.970&.959&.755&8.38\\
 & RSE$\downarrow$ & .499&.686&.817&.881&.397&.534&.565&.580&.210&.431&.621&.601&.271&.303&.326&.481&.513&7.81\\
\midrule
iTransformer & $R^2\uparrow$ & .829&\textbf{.623}&.439&.285&.887&.719&.685&.668&\textbf{.964}&.879&.799&.738&.979&.977&.975&.964&.776&4.84\\
 & RSE$\downarrow$ & .436&\textbf{.648}&.780&.878&.362&.547&\underline{.561}&.584&\underline{.191}&.348&.448&.563&.259&.305&.335&.427&.479&5.09\\
\midrule
TS-TCN & $R^2\uparrow$ & .810&.605&\textbf{.473}&\underline{.328}&\textbf{.897}&\underline{.759}&\underline{.698}&.652&\textbf{.964}&.884&.762&.720&.980&.971&.968&.962&.777&5.41\\
 & RSE$\downarrow$ & .459&.656&\underline{.757}&\underline{.857}&\textbf{.354}&\underline{.527}&\textbf{.559}&.633&\textbf{.189}&\textbf{.325}&.484&.523&.264&.316&.318&.360&.474&4.22\\
\midrule
TS-GRU & $R^2\uparrow$ & \textbf{.848}&.618&.430&\textbf{.329}&.874&.742&.684&.649&.938&.878&.779&.722&\underline{.991}&.981&\underline{.983}&.976&.776&5.47\\
 & RSE$\downarrow$ & \textbf{.412}&.651&.795&\textbf{.853}&.384&.530&.587&.637&.253&.349&\underline{.426}&.527&.216&.240&.236&.271&.460&5.19\\
\midrule
iSpikformer & $R^2\uparrow$ & .817&.618&.440&.279&.879&.744&.687&\textbf{.674}&.961&.876&.795&.738&.977&.974&.972&.963&.775&5.88\\
 & RSE$\downarrow$ & .475&.668&\textbf{.752}&.905&.376&.536&.569&.580&.204&.333&.465&.521&.263&.284&.338&.348&.476&5.56\\
\midrule
TS-former & $R^2\uparrow$ & \underline{.847}&.620&.445&.283&.874&.735&.683&.669&.961&\textbf{.886}&\textbf{.828}&\textbf{.774}&.987&\underline{.985}&.981&\underline{.977}&\underline{.783}&\underline{3.78}\\
 & RSE$\downarrow$ & .416&.655&.783&.874&.379&.539&.572&.583&.224&\underline{.331}&\textbf{.382}&\textbf{.435}&\underline{.197}&\underline{.215}&\underline{.234}&\underline{.261}&\underline{.443}&\underline{4.16}\\
\midrule
\textbf{SpikeLite} & $R^2\uparrow$ & \underline{.847}&\underline{.622}&\underline{.452}&.316&\underline{.889}&\textbf{.781}&\textbf{.715}&.639&\underline{.962}&\underline{.885}&\underline{.811}&\underline{.767}&\textbf{.993}&\textbf{.991}&\textbf{.988}&\textbf{.983}&\textbf{.790}&\textbf{2.38}\\
 & RSE$\downarrow$ & \underline{.414}&\underline{.649}&.781&.873&\underline{.361}&\textbf{.506}&.577&.649&.201&.348&.446&\underline{.496}&\textbf{.146}&\textbf{.167}&\textbf{.196}&\textbf{.233}&\textbf{.440}&\textbf{3.38}\\
\bottomrule
\end{tabular}}
\normalsize
\end{table*}

\paragraph{Compared methods.}
For the SeqSNN protocol, we compare against statistical predictors, including ARIMA and Gaussian Process regression, and ANN forecasters, including Autoformer, PatchTST, and iTransformer.
ARIMA, Gaussian Process, Autoformer, iTransformer, and iSpikformer results are taken from SeqSNN~\cite{lv2024efficient}.
TS-TCN, TS-GRU, and TS-former results are taken from TS-LIF~\cite{feng2025tslif}.
PatchTST, AGCRN, and STAEformer are implemented and evaluated under the SeqSNN data splits and evaluation procedure, using their original architectures~\cite{nie2023patchtst,bai2020agcrn,liu2023staeformer}.

For the SpikF protocol (long-term forecasting), we compare SpikeLite with the ANN baselines iTransformer, RLinear, PatchTST, Crossformer, TimesNet, DLinear, SCINet,
and Autoformer, together with SpikF.
Their reported results are taken from the SpikF benchmark~\cite{wu2025spikf}, while SpikeLite is evaluated under the same protocol.

\subsection{Main Results}
\label{sec:main_results}

\begin{table*}[t]
\centering
\caption{
Long-term forecasting performance under the SpikF protocol~\cite{wu2025spikf}.
Results are averaged across four prediction horizons ($96$, $192$, $336$, and $720$) and three random seeds.
Lower MSE and MAE indicate better performance.
\textbf{Bold} and \underline{underlined} values indicate the best and second-best results, respectively.
}
\label{tab:spikf_average}
\scriptsize
\setlength{\tabcolsep}{1.6pt}
\renewcommand{\arraystretch}{1.04}
\resizebox{\linewidth}{!}{%
\begin{tabular}{@{}l*{10}{cc}@{}}
\toprule
Dataset & \multicolumn{2}{c}{iTransformer} & \multicolumn{2}{c}{RLinear} & \multicolumn{2}{c}{PatchTST} & \multicolumn{2}{c}{Crossformer} & \multicolumn{2}{c}{TimesNet} & \multicolumn{2}{c}{DLinear} & \multicolumn{2}{c}{SCINet} & \multicolumn{2}{c}{Autoformer} & \multicolumn{2}{c}{\textbf{SpikF}} & \multicolumn{2}{c}{\textbf{SpikeLite}}\\
\cmidrule(lr){2-3}\cmidrule(lr){4-5}\cmidrule(lr){6-7}\cmidrule(lr){8-9}\cmidrule(lr){10-11}\cmidrule(lr){12-13}\cmidrule(lr){14-15}\cmidrule(lr){16-17}\cmidrule(lr){18-19}\cmidrule(l){20-21}
 & MSE & MAE & MSE & MAE & MSE & MAE & MSE & MAE & MSE & MAE & MSE & MAE & MSE & MAE & MSE & MAE & MSE & MAE & MSE & MAE\\
\midrule
ECL      & \underline{.178} & \underline{.270} & .219 & .298 & .205 & .290 & .244 & .334 & .192 & .295 & .212 & .300 & .268 & .365 & .227 & .338 & .183 & .275 & \textbf{.175} & \textbf{.263}\\
Weather  & \underline{.258} & \underline{.278} & .272 & .291 & .259 & .281 & .259 & .315 & .259 & .287 & .265 & .317 & .292 & .363 & .338 & .382 & \textbf{.245} & \textbf{.265} & \textbf{.245} & \textbf{.265}\\
ETTh1    & .454 & .447 & .446 & .434 & .469 & .454 & .529 & .522 & .458 & .450 & .456 & .452 & .747 & .647 & .496 & .487 & \textbf{.440} & \textbf{.428} & \underline{.443} & \underline{.429}\\
ETTh2    & .383 & .407 & \underline{.374} & \underline{.398} & .387 & .407 & .942 & .684 & .414 & .427 & .559 & .515 & .954 & .723 & .450 & .459 & \textbf{.372} & \textbf{.394} & \underline{.374} & \textbf{.394}\\
ETTm1    & .407 & .410 & .414 & .407 & \textbf{.387} & .400 & .513 & .496 & .400 & .406 & .403 & .407 & .485 & .481 & .588 & .517 & \underline{.388} & \textbf{.385} & \textbf{.387} & \underline{.389}\\
ETTm2    & .288 & .332 & .286 & .327 & \underline{.281} & .326 & .757 & .610 & .291 & .333 & .350 & .401 & .571 & .537 & .327 & .371 & \underline{.281} & \underline{.320} & \textbf{.278} & \textbf{.319}\\
Traffic  & \textbf{.428} & \textbf{.282} & .626 & .378 & \underline{.481} & .304 & .550 & .304 & .620 & .336 & .625 & .383 & .804 & .509 & .628 & .379 & .497 & .296 & \underline{.479} & \underline{.294}\\
Exchange & \underline{.360} & \underline{.403} & \textbf{.354} & .414 & .367 & .404 & .940 & .707 & .416 & .443 & \textbf{.354} & .414 & .750 & .626 & .613 & .539 & \underline{.360} & \textbf{.402} & .366 & .404\\
\midrule
Avg. & \underline{.345} & .354 & .374 & .368 & .355 & .358 & .592 & .497 & .381 & .372 & .403 & .399 & .609 & .531 & .458 & .434 & .346 & \underline{.346} & \textbf{.343} & \textbf{.345}\\
Avg. Rank & 3.56 & 3.56 & 5.12 & 5.00 & 4.38 & 4.19 & 8.25 & 8.31 & 5.50 & 5.50 & 5.94 & 7.12 & 9.38 & 9.38 & 8.38 & 8.38 & \underline{2.44} & \textbf{1.75} & \textbf{2.06} & \underline{1.81}\\
\bottomrule
\end{tabular}}
\normalsize
\end{table*}

SpikeLite achieves the best overall forecasting performance under both the standard multivariate and long-term forecasting settings.

Under the \textbf{SeqSNN protocol}, SpikeLite achieves the best overall performance, with an average $R^2$ of $0.790$ and an average RSE of $0.440$. Across the 16 dataset--horizon settings for each metric, SpikeLite obtains the largest number of first- and second-place results.

Among the four benchmarks, the clearest improvement appears on Electricity, whose pronounced periodic structure is well aligned with FSSE's reorganization of the input into multi-scale frequency-sensitive components. SpikeLite also performs strongly on METR-LA and PEMS-BAY, where correlated sensor measurements require both temporal modeling and information exchange across variables. 
Among the compared methods, TS-former ranks second and shows a slight advantage over SpikeLite on Solar, possibly due to its stronger emphasis on local temporal dynamics.

Under the \textbf{SpikF protocol}, SpikeLite achieves the best average MSE and MAE, demonstrating that its performance advantage also holds under the long-term forecasting setting.

SpikeLite extends a consistent advantage on Electricity to ECL, substantially outperforming all compared methods in long-term forecasting.
Across the ETT and Traffic benchmarks, SpikeLite remains competitive with the strongest ANN and SNN baselines. The ETT datasets exhibit multi-scale temperature and load variations, whereas Traffic contains recurring patterns among spatially correlated sensor measurements; FSSE and SSCA jointly model these temporal and cross-channel characteristics. Exchange is more challenging due to its less regular long-horizon variations, yet SpikeLite remains close to the best-performing methods.

\subsection{Frequency-Selective Mechanism Analysis}
\label{sec:frequency_analysis}

  \begin{wrapfigure}{r}{0.45\linewidth}
      \vspace{-6pt}
      \centering
      \includegraphics[width=\linewidth]{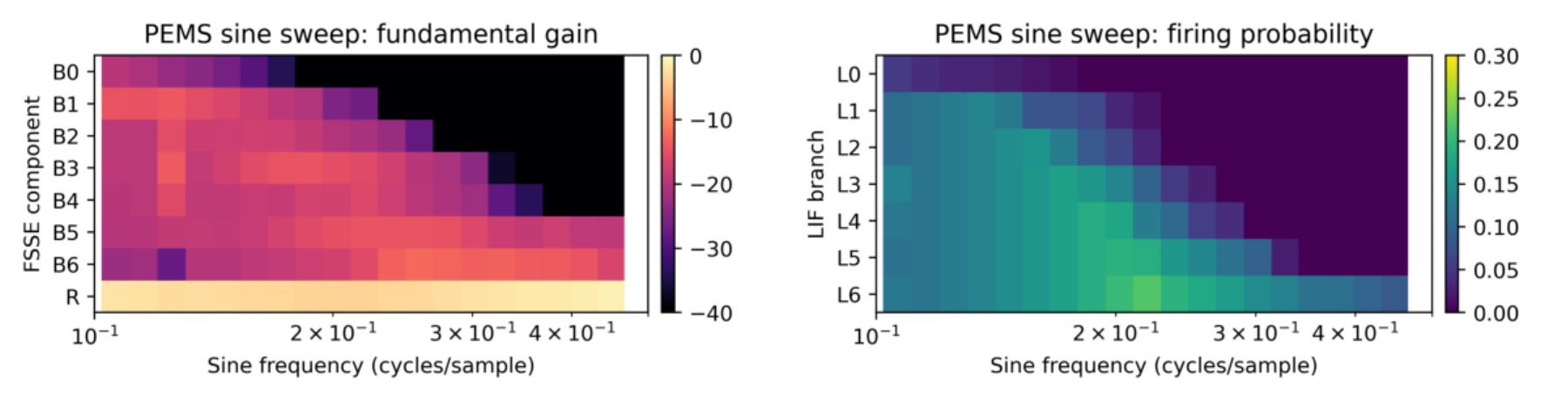}
      \caption{Frequency-sweep analysis of FSSE: component-wise fundamental gain and branch-wise firing probability.}
      \label{fig:frequency_mechanism}
      \vspace{-8pt}
  \end{wrapfigure}

We further examine whether the parallel LIF branches and the resulting FSSE components exhibit distinct responses to temporal frequencies. 
Following the frequency-response analysis in Section~\ref{sec:fsse}, we apply sinusoidal inputs to the PEMS setting and sweep the frequency from $0.1$ to $0.5$ cycles per sample. 
The higher-frequency portion of this sweep provides a clearer view of the differences among the learned responses, as shown in Fig.~\ref{fig:frequency_mechanism}.

In Fig.~\ref{fig:frequency_mechanism}(a), the FSSE components exhibit distinct gain profiles: earlier components respond more strongly at lower frequencies, whereas later components and the residual remain responsive over a broader frequency range. 
Fig.~\ref{fig:frequency_mechanism}(b) shows a similar overall trend in firing probability. 
The firing-probability profiles broadly follow the nominal slow-to-fast initialization order, while allowing channel-dependent deviations after training.
These results are consistent with the heterogeneous decay factors learned by FSSE and demonstrate that heterogeneous LIF dynamics and event gating produce distinct frequency-sensitive responses.

\subsection{Ablation Study}
\label{sec:ablation}

  \begin{wrapfigure}{r}{0.46\linewidth}
      \vspace{-6pt}
      \centering
      \includegraphics[width=\linewidth]{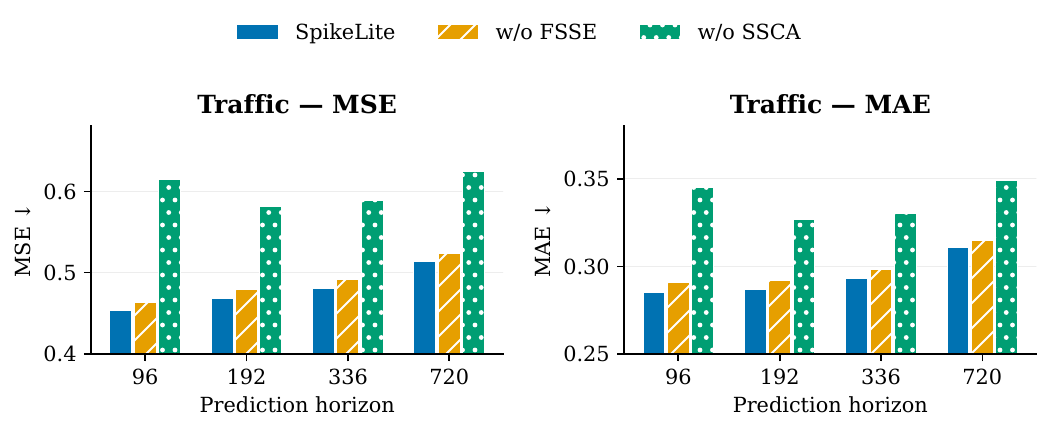}
      \caption{Ablation results on Traffic under the SpikF protocol. Lower values indicate better forecasting performance.}
      \label{fig:ablation_traffic}
      \vspace{-8pt}
  \end{wrapfigure}

To assess the contributions of FSSE and SSCA, we perform an ablation study under the SpikF protocol. The main-text analysis focuses on Traffic, where dependencies among observed variables make cross-channel modeling particularly relevant. The full model is compared against \textit{w/o FSSE} and \textit{w/o SSCA} with all other components and training settings unchanged. Figure~\ref{fig:ablation_traffic} reports the Traffic results, with the corresponding Weather results provided in Appendix~\ref{app:detailed_ablations}.

Removing either component degrades performance across all four Traffic horizons.
Without FSSE, the average MSE and MAE increase from $0.479$ and $0.294$ to $0.489$ and $0.299$, indicating that frequency-sensitive encoding improves the representation of each variable's temporal variations.
Removing SSCA produces a larger increase, to $0.602$ MSE and $0.338$ MAE, showing that selective cross-channel interaction contributes substantial predictive information on Traffic. 
Together, these results support the complementary roles of FSSE in temporal encoding and SSCA in modeling dependencies among variables.

\subsection{Energy Analysis}
\label{sec:energy}

Following the established energy-estimation protocol based on 45-nm CMOS technology~\cite{horowitz20141}, where a multiply--accumulate operation is assigned an energy cost of $4.6$ pJ and an accumulate operation is assigned $0.9$ pJ, we estimate the inference cost of the encoder backbone. 
The task-specific prediction heads are excluded for all methods to ensure a consistent comparison.

\begin{wraptable}{r}{0.45\linewidth}
    \vspace{-6pt}
    \centering
    \caption{
    Model complexity, estimated energy, and forecasting accuracy on ECL at a prediction horizon of 720. 
    }
    \label{tab:energy_comparison}
    \setlength{\tabcolsep}{2.8pt}
    \renewcommand{\arraystretch}{1.05}
    \scriptsize
    \begin{tabular}{lcccc}
    \toprule
    Model & Params & OPs & Energy ($\mu$J) & MSE@720 \\
    \midrule
    SpikF        & 1.2K  & 0.13G & 117.66 & .219 \\
    DLinear      & 0.14M & 45M   & 205.19 & .245 \\
    iTransformer & 1.6M  & 0.72G & 3289.29 & .225 \\
    \midrule
    SpikeLite    & 67.2K & 0.02G & 95.30 & .218 \\
    \bottomrule
    \end{tabular}
    \vspace{-8pt}
\end{wraptable}

Table~\ref{tab:energy_comparison} shows that SpikeLite uses only $67.2$K parameters and $0.02$G operations, corresponding to an estimated energy consumption of $95.30\,\mu$J per sample, which is the lowest among the compared models. 
Compared with SpikF, SpikeLite reduces energy consumption by approximately $19.0\%$, while the savings over the ANN-based baselines are approximately $53.6\%$ relative to DLinear and $97.1\%$ relative to iTransformer.

These results indicate that the energy advantage of SpikeLite is attributed to its spike-driven encoder and compact channel interaction pathway, which reduce the number of costly operations required during inference while maintaining forecasting accuracy.

\section{Conclusion}

We presented SpikeLite, a lightweight spiking framework for multivariate time-series forecasting.
FSSE uses heterogeneous LIF dynamics to reorganize each input sequence into frequency-sensitive components, while SSCA selectively models informative cross-channel interactions through a sparse binary mask.
Experiments under the SeqSNN and SpikF protocols show that SpikeLite achieves consistently strong forecasting accuracy across standard and long-term benchmarks, while requiring the lowest estimated energy among the compared models.
The limitations and future directions are discussed in Appendix~\ref{app:limitations_future}.

\section{Reproducibility Statement}
\label{sec:reproducibility}

The supplementary material documents the dataset characteristics, temporal properties, preprocessing and evaluation protocols, and implementation and training configurations (Appendix~\ref{app:experimental_details}).
It also provides the complete horizon-wise results, seed-stability analysis, ablations, frequency diagnostics, and energy-accounting details. 
Results produced by our implementation are averaged over three independent random seeds unless stated otherwise.
All corresponding mean$\pm$std results are reported in Appendix~\ref{app:robustness}.
Numerical results copied from published benchmark reports are identified in the corresponding comparison description and are retained under their original evaluation protocol; they are not presented as new three-seed measurements.
The source code, configuration files, and analysis scripts will be released upon publication.

\section{Ethics Statement}
\label{sec:ethics}

Our experiments use publicly available multivariate time-series benchmarks covering traffic sensors, electricity and solar measurements, weather observations, transformer measurements, and exchange rates. 
We collect no new human or animal data, and our experiments do not use individual-level identifiers or attempt to infer personal attributes. 
The datasets are used for research in accordance with their original licenses and documentation.
Although aggregate traffic, energy, and exchange rates can support sensitive operational or mobility-related inferences in downstream systems, such deployments should follow the applicable data-use restrictions, privacy requirements, and access-control practices.

\bibliographystyle{unsrt}  
\bibliography{references}

\appendix


\section{Experimental Details}
\label{app:experimental_details}

This appendix records the dataset characteristics, evaluation rules,
implementation choices, additional forecasting results, and diagnostic
experiments used in the paper. The two protocols are kept separate because
they use different datasets, lookback windows, horizons, metrics, and model
interfaces.

\subsection{Datasets and temporal characteristics}
\label{app:dataset_details}

Table~\ref{tab:dataset_statistics} lists the dimensionality exposed to the
forecasting model. The sensor-network datasets contain substantial
cross-variable structure, whereas the physical and financial datasets are
more heterogeneous in their temporal regularity. All datasets used here are
regularly sampled benchmark series with complete input windows; explicit
missing-value handling is not part of the current pipeline.

\begin{table}[H]
\centering
\caption{Datasets used by the two evaluation protocols. ``Variables/nodes''
denotes the input dimensionality after protocol preprocessing.}
\label{tab:dataset_statistics}
\scriptsize
\setlength{\tabcolsep}{3.5pt}
\renewcommand{\arraystretch}{1.08}
\begin{tabular}{@{}llllp{5.4cm}@{}}
\toprule
Protocol & Dataset & Domain & Variables/nodes & Main temporal characteristics \\
\midrule
SeqSNN & METR-LA & traffic sensors & 207 & Road-speed measurements with spatial correlation, daily repetition, and short-term congestion changes. \\
       & PEMS-BAY & traffic sensors & 325 & Large sensor network with correlated locations, recurring traffic cycles, and rapidly changing local conditions. \\
       & Solar & solar power & 137 & Diurnal structure with weather-driven local fluctuations and intermittent high-frequency changes. \\
       & Electricity & electricity load & 321 & Heterogeneous consumption channels with periodic structure and channel-specific temporal profiles. \\
\midrule
SpikF & ECL & electricity load & 321 & Long-horizon electricity demand with periodic and heterogeneous channel behavior. \\
      & Weather & meteorological series & 21 & Smooth physical variables with seasonal trends and multi-scale local variation. \\
      & ETTh1/ETTh2 & transformer temperature & 7 & Hourly transformer measurements combining trend, periodicity, and regime-dependent variation. \\
      & ETTm1/ETTm2 & transformer temperature & 7 & Higher-frequency ETT benchmarks with finer local fluctuations. \\
      & Traffic & road sensors & 862 & High-dimensional traffic network with recurring patterns and dense cross-sensor dependencies. \\
      & Exchange & exchange rates & 8 & Financial series with weaker periodicity, nonstationary long-range behavior, and less stable correlations. \\
\bottomrule
\end{tabular}
\end{table}

METR-LA and PEMS-BAY are traffic-network settings in the standard
multivariate protocol; their spatial correlations make cross-channel
information potentially useful. Solar and Electricity are dominated by
periodic energy-generation or consumption patterns, but individual channels
still exhibit different amplitudes and local variations. In the long-term
protocol, ECL repeats the electricity domain with a distinct preprocessing
pipeline, while the ETT datasets expose trend and periodic components at
different sampling frequencies. Traffic stresses both long-range temporal
structure and channel interaction. Exchange is a contrasting case: its
irregular and weakly periodic behavior makes small absolute metric
differences produce noticeable ranking changes.

\subsection{Evaluation protocols and preprocessing}
\label{app:protocols}

Under the standard multivariate protocol, we evaluate METR-LA, PEMS-BAY,
Solar, and Electricity at horizons $\{6,24,48,96\}$ using $R^2$ and RSE.
The released SeqSNN data splits, normalization, and dataset-specific input
windows are retained. The long-term protocol uses a lookback window of $96$
and horizons $\{96,192,336,720\}$, evaluated with MSE and MAE. The complete
horizon-wise table is given in Section~\ref{app:full_ltsf_results}.

\subsection{Evaluation metrics}
\label{app:metrics}

For the standard multivariate benchmarks, the four datasets have different
scales, dimensionalities, and horizon lengths. We therefore use the global
scale-normalized metrics adopted by the protocol. With ground truth $y_i$,
prediction $\widehat y_i$, and global mean $\bar y$, they are
\begin{equation}
R^2=1-\frac{\sum_i(y_i-\widehat y_i)^2}{\sum_i(y_i-\bar y)^2},
\qquad
\mathrm{RSE}=\frac{\sqrt{\sum_i(y_i-\widehat y_i)^2}}
{\sqrt{\sum_i(y_i-\bar y)^2}}.
\label{eq:app_seq_metrics}
\end{equation}
This normalization makes errors comparable across datasets whose raw units
and variances differ. Higher $R^2$ and lower RSE are better.

For long-term forecasting, we report
\begin{equation}
\mathrm{MSE}=\frac{1}{N}\sum_{i=1}^{N}(y_i-\widehat y_i)^2,
\qquad
\mathrm{MAE}=\frac{1}{N}\sum_{i=1}^{N}|y_i-\widehat y_i|,
\label{eq:app_ltsf_metrics}
\end{equation}
where $N$ counts all forecast entries after flattening the horizon and
channel dimensions. Dataset averages are computed after averaging the four
horizons within each dataset.

\subsection{Implementation and reproducibility}
\label{app:implementation}

The SeqSNN configurations use $D=128$ for the compact channel
representation. The released ``spikelinear'' configurations use a window of
$12$ for PEMS-BAY and dataset-specific windows for the remaining standard
benchmarks. FSSE uses per-channel learnable decay factors and thresholds,
surrogate-gradient scale $4$, RevIN, component normalization, and horizon
scaling. SSCA uses four attention heads, $T_s=4$ simulation steps, query--key
scale $0.125$, mask rank $8$, $\gamma=1$, and mask threshold $0.3$; stochastic
mask sampling is disabled in both training and evaluation.

We use Adam with zero weight decay, batch size $32$, learning rates specified
by the released configurations, and early stopping on validation loss. Main
table entries follow the configured repeated runs

\section{Complete Long-Term Forecasting Results}
\label{app:full_ltsf_results}

Table~\ref{tab:spikf_full} reports all dataset--horizon combinations under
the long-term protocol. Bold and underlined values denote the best and
second-best entries, respectively.

\input{spikf_full_table.tex}

The horizon-wise results reveal differences that are obscured by the
dataset averages. On ECL, SpikeLite obtains the lowest MSE and MAE at all
four horizons, indicating that its advantage is maintained as the
prediction length increases. On Weather, its average results are
comparable to SpikF, with small differences that change direction across
horizons. Among the ETT datasets, SpikeLite achieves the lowest average
MSE and MAE on ETTm2, whereas the results on ETTh1, ETTh2, and ETTm1 are
close to those of the strongest competing methods rather than uniformly
better at every horizon.

The table also clarifies where SpikeLite is less competitive. On Traffic,
it improves on SpikF at every horizon but trails iTransformer, which
obtains the lowest errors on this dataset. On Exchange, the gap to SpikF
is small at shorter horizons, while the larger error at horizon $720$
raises SpikeLite's dataset average. These cases qualify the aggregate
results in the main text: the overall performance does not imply
consistent superiority on every dataset or prediction horizon.
\section{Stability Across Random Seeds}
\label{app:robustness}

To assess the robustness of SpikeLite to random initialization, we repeat
the experiments with three independent random seeds under both the standard
multivariate and long-term forecasting protocols. For a metric value
$q\in\{R^2,\mathrm{RSE},\mathrm{MSE},\mathrm{MAE}\}$, we report the sample
mean and standard deviation across the three runs:
\begin{equation}
s(q)=
\sqrt{
\frac{1}{N-1}
\sum_{n=1}^{N}
\left(q_n-\bar q\right)^2
},
\qquad N=3.
\label{eq:app_std}
\end{equation}

Table~\ref{tab:three_seed_ltsf} reports the three-seed results under the
SpikF protocol. The ``Avg.'' column is first computed within each run by
averaging the four prediction horizons, after which the mean and standard
deviation are calculated across the three runs. Table~\ref{tab:robustness}
provides the corresponding results under the standard multivariate protocol.

\begin{table*}[t]
\centering
\caption{Three-seed stability under the SpikF protocol. Each entry reports
the mean $\pm$ sample standard deviation over three independent runs. The
Avg. column is computed within each run over the four prediction horizons.}
\label{tab:three_seed_ltsf}
\scriptsize
\setlength{\tabcolsep}{3.0pt}
\renewcommand{\arraystretch}{1.05}
\begin{tabular}{@{}llccccc@{}}
\toprule
Dataset & Metric & 96 & 192 & 336 & 720 & Avg. \\
\midrule
\multirow{2}{*}{ECL}
& MSE & $.1453\!\pm\!.0006$ & $.1610\!\pm\!.0000$ & $.1750\!\pm\!.0000$ & $.2187\!\pm\!.0012$ & $.1750\!\pm\!.0003$ \\
& MAE & $.2363\!\pm\!.0006$ & $.2500\!\pm\!.0000$ & $.2640\!\pm\!.0000$ & $.3007\!\pm\!.0012$ & $.2628\!\pm\!.0003$ \\
\midrule
\multirow{2}{*}{Weather}
& MSE & $.1587\!\pm\!.0012$ & $.2090\!\pm\!.0000$ & $.2660\!\pm\!.0000$ & $.3460\!\pm\!.0000$ & $.2449\!\pm\!.0003$ \\
& MAE & $.1977\!\pm\!.0012$ & $.2440\!\pm\!.0000$ & $.2853\!\pm\!.0006$ & $.3380\!\pm\!.0000$ & $.2663\!\pm\!.0003$ \\
\midrule
\multirow{2}{*}{ETTm2}
& MSE & $.1753\!\pm\!.0006$ & $.2400\!\pm\!.0000$ & $.3023\!\pm\!.0006$ & $.3950\!\pm\!.0000$ & $.2782\!\pm\!.0003$ \\
& MAE & $.2540\!\pm\!.0000$ & $.2950\!\pm\!.0000$ & $.3360\!\pm\!.0000$ & $.3900\!\pm\!.0000$ & $.3188\!\pm\!.0000$ \\
\midrule
\multirow{2}{*}{Traffic}
& MSE & $.4530\!\pm\!.0000$ & $.4680\!\pm\!.0000$ & $.4810\!\pm\!.0000$ & $.5140\!\pm\!.0000$ & $.4790\!\pm\!.0000$ \\
& MAE & $.2850\!\pm\!.0000$ & $.2870\!\pm\!.0000$ & $.2930\!\pm\!.0000$ & $.3110\!\pm\!.0000$ & $.2940\!\pm\!.0000$ \\
\bottomrule
\end{tabular}
\end{table*}

\begin{table}[h]
\centering
\small
\caption{Three-seed stability under the standard multivariate protocol.
Each entry reports the mean $\pm$ sample standard deviation over three
independent runs.}
\label{tab:robustness}
\begin{tabular}{l l c c c c}
\toprule
\textbf{Dataset} & \textbf{Metric} & \textbf{Horizon 6} & \textbf{Horizon 24} & \textbf{Horizon 48} & \textbf{Horizon 96} \\
\midrule
\multirow{2}{*}{PEMS-BAY}
 & $R^2\!\uparrow$  & $.899 \pm .001$ & $.781 \pm .001$ & $.715 \pm .000$ & $.639 \pm .003$ \\
 & $RSE\downarrow$   & $.361 \pm .000$ & $.506 \pm .001$ & $.577 \pm .000$ & $.649 \pm .002$ \\
\midrule
\multirow{2}{*}{METR-LA}
 & $R^2\!\uparrow$  & $.847 \pm .001$ & $.622 \pm .000$ & $.452 \pm .000$ & $.316 \pm .001$ \\
 & $RSE\downarrow$   & $.414 \pm .001$ & $.649 \pm .000$ & $.781 \pm .002$ & $.873 \pm .002$ \\
\midrule
\multirow{2}{*}{Solar}
 & $R^2\!\uparrow$  & $.962 \pm .000$ & $.885 \pm .001$ & $.810 \pm .001$ & $.767 \pm .001$ \\
 & $RSE\downarrow$   & $.201 \pm .000$ & $.348 \pm .000$ & $.446 \pm .000$ & $.496 \pm .000$ \\
\midrule
\multirow{2}{*}{Electricity}
 & $R^2\!\uparrow$  & $.993 \pm .000$ & $.991 \pm .001$ & $.988 \pm .000$ & $.983 \pm .000$ \\
 & $RSE\downarrow$   & $.146 \pm .000$ & $.167 \pm .001$ & $.196 \pm .000$ & $.233 \pm .000$ \\
\bottomrule
\end{tabular}
\end{table}

The results show that SpikeLite is stable across random initializations.
Under the SpikF protocol, the standard deviations are at most $0.0012$ for
both MSE and MAE, and several entries remain identical after rounding.
Under the standard multivariate protocol, most standard deviations are at
most $0.001$, with the largest observed deviation equal to $0.003$.
Therefore, the main performance trends and relative rankings are not
sensitive to the particular random seed.

\section{Additional Module Ablations}
\label{app:detailed_ablations}

The main text focuses on Traffic because its sensor variables exhibit strong
useful interactions. Table~\ref{tab:traffic_weather_ablation} gives the
horizon-wise Traffic and Weather results for removing FSSE or SSCA while
keeping all other settings fixed.

\begin{table}[H]
\centering
\caption{Horizon-wise ablations under the long-term protocol. Lower MSE and
MAE are better.}
\label{tab:traffic_weather_ablation}
\scriptsize
\setlength{\tabcolsep}{3pt}
\renewcommand{\arraystretch}{1.05}
\begin{tabular}{@{}ll*{8}{c}@{}}
\toprule
Dataset & Variant & \multicolumn{2}{c}{96} & \multicolumn{2}{c}{192}
& \multicolumn{2}{c}{336} & \multicolumn{2}{c}{720} \\
\cmidrule(lr){3-4}\cmidrule(lr){5-6}\cmidrule(lr){7-8}\cmidrule(l){9-10}
& & MSE & MAE & MSE & MAE & MSE & MAE & MSE & MAE \\
\midrule
\multirow{3}{*}{Traffic}
& SpikeLite & .453 & .285 & .468 & .287 & .481 & .293 & .514 & .311 \\
& w/o FSSE & .463 & .291 & .479 & .292 & .491 & .298 & .524 & .315 \\
& w/o SSCA & .615 & .345 & .581 & .327 & .589 & .330 & .624 & .349 \\
\midrule
\multirow{3}{*}{Weather}
& SpikeLite & .158 & .197 & .209 & .244 & .266 & .285 & .346 & .338 \\
& w/o FSSE & .172 & .210 & .223 & .256 & .278 & .297 & .358 & .350 \\
& w/o SSCA & .179 & .215 & .224 & .256 & .278 & .296 & .356 & .357 \\
\bottomrule
\end{tabular}
\end{table}

On Traffic, removing FSSE raises the averaged MSE/MAE from $(.479,.294)$ to
$(.489,.299)$, while removing SSCA produces the larger degradation to
$(.602,.338)$. Weather shows smaller module gaps, consistent with weaker
benefits from explicit cross-channel exchange. These results support the
conditional design: FSSE supplies the input representation, while SSCA is
most useful when channel interaction carries predictive information.

\section{Frequency-Selective Mechanism Diagnostics}
\label{app:frequency_diagnostics}

We analyze FSSE before its temporal projections and before SSCA. For a
component $b_{n,c}^{(j)}[l]$ from test window $n$ and channel $c$, we remove
the temporal mean, apply a Hann window $w[l]$, and compute
\begin{equation}
P_{n,c}^{(j)}(f)=\left|\operatorname{rFFT}\left(w[l]
\left(b_{n,c}^{(j)}[l]-\bar b_{n,c}^{(j)}\right)\right)\right|^2.
\label{eq:app_periodogram}
\end{equation}
The spectral centroid used in the real-data summary is
\begin{equation}
\mu_f^{(j)}=\frac{\sum_{f>0}fP^{(j)}(f)}{\sum_{f>0}P^{(j)}(f)}.
\label{eq:app_centroid}
\end{equation}

\begin{figure*}[t]
\centering
\includegraphics[width=.80\linewidth]{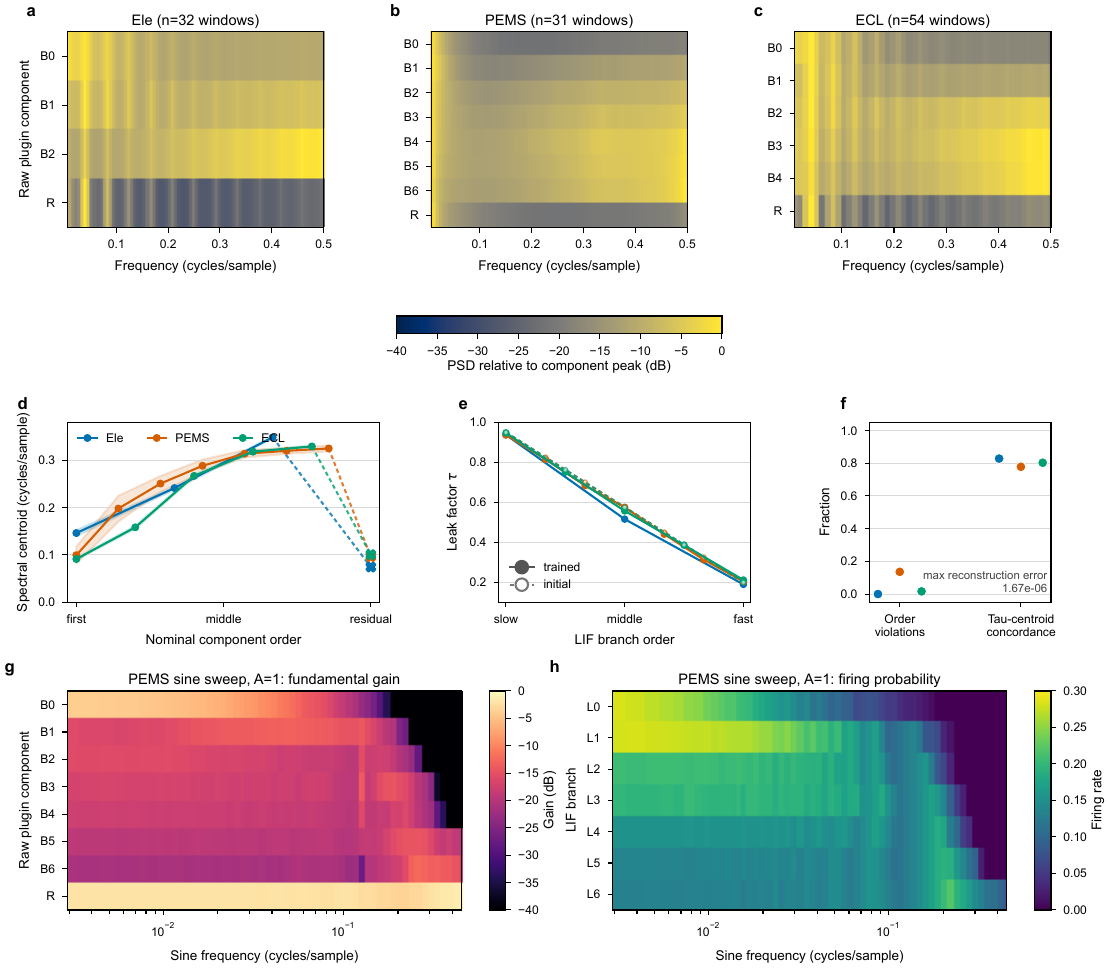}
\caption{Frequency diagnostics for FSSE. Panels (a)--(c): real-data
component spectra. Panel (d): spectral centroids. Panel (e): initialized and
trained decay factors. Panel (f): branch-order violations and decay--centroid
concordance. Panels (g)--(h): PEMS sinusoidal fundamental gain and firing
probability.}
\label{fig:app_frequency_diagnostics}
\end{figure*}

Figure~\ref{fig:app_frequency_diagnostics} makes the panel mapping explicit:
panels (a)--(c) show component spectra for Electricity, PEMS-BAY, and ECL;
(d) compares component spectral centroids; (e) compares initialized and
trained decay factors; (f) reports order violations and decay--centroid
concordance; and (g)--(h) show the PEMS sine-sweep fundamental gain and LIF
firing probability. The branches are frequency-sensitive but are not forced
into disjoint bands, which is consistent with the nonlinear threshold, reset,
and event-gating operations.

Before learnable component scaling and temporal projection, the successive
difference construction is exactly reconstructive:
$
\sum_{j=0}^{K}\mathbf{B}^{(j)}=\mathbf{X}.
\label{eq:app_reconstruction}
$.
The largest floating-point reconstruction error in the diagnostic is
$1.67\times10^{-6}$. Thus, FSSE reorganizes the input response without
discarding the signal at the decomposition stage.

\section{Energy Accounting}
\label{app:energy_accounting}

We report the encoder-level energy estimate for the ECL dataset with a
prediction horizon of 720. We follow the established 45-nm CMOS accounting
convention, assigning $E_{\mathrm{MAC}}=4.6$ pJ to a dense
multiply--accumulate operation and $E_{\mathrm{AC}}=0.9$ pJ to an
accumulation. The task-specific prediction head is excluded for every model
in the comparison.

For SpikeLite, the dense operation count is decomposed into the FSSE
encoder, temporal projections, and, when enabled, SSCA:
\begin{align}
N_{\mathrm{dense}}={}&3KLC+(K+1)CLD \\
&+\mathbb{I}_{\mathrm{SSCA}}
\left[
2T_sCD+
T_s(3CD^2+2C^2D+CD^2)
+N_{\mathrm{mask}}
\right],
\label{eq:app_ops}
\end{align}
where $N_{\mathrm{mask}}$ accounts for the rFFT and low-rank affinity
calculation. Measured spike rates are used to convert spike-driven terms
into effective accumulation operations. The final energy estimate is
\begin{equation}
E=
E_{\mathrm{MAC}}N_{\mathrm{MAC}}
+
E_{\mathrm{AC}}N_{\mathrm{AC}}.
\label{eq:app_energy}
\end{equation}

\section{Toward Hardware Realization}
\label{app:hardware_realization}

SpikeLite uses hardware-compatible primitives, but the following discussion is
an implementability analysis rather than a measured hardware result.

\paragraph{FSSE dataflow.}
Each input channel and LIF branch maintains one membrane state and one
previous-event state. Leaky integration, thresholding, subtractive reset, and
event gating are local operations with no all-to-all communication. Successive
differences and the terminal residual require only additions and subtractions.
Inactive events can be omitted from a compressed event stream, while an
active event carries its membrane payload; this maps the encoder to a mixed event-control and membrane-value data path.

\paragraph{SSCA dataflow.}
SSCA can tile the Q/K/V projections over channels and hidden dimensions. The
binary mask is generated once per input window and stored as a bit matrix or
index list. Masked channel blocks can then be skipped by a scheduler during
the $T_s$ simulation steps. For large $C$, blockwise generation avoids
materializing a dense score matrix; for dense masks, the implementation can
fall back to regular tiled attention.

\section{Limitations and Future Work}
\label{app:limitations_future}

\paragraph{Limitations.}
The current benchmarks provide regularly sampled, fully observed windows, so
SpikeLite does not yet model missing values, irregular timestamps, or online
arrival of observations. The reported results therefore establish the method
under complete-window forecasting rather than under data-imputation or
irregular-sampling conditions.

\paragraph{Future work.}
We will implement FSSE and SSCA on programmable or fabricated neuromorphic
hardware and measure actual energy, latency, memory traffic, and sparse
execution efficiency. We will also study irregular time series,
missing-data forecasting, online prediction, and cross-dataset transfer in the future work.

\end{document}

%% file: spikf_full_table.tex
\begin{table*}[t]
\centering
\caption{Full horizon-wise long-term forecasting results under the SpikF
protocol. Each entry reports MSE and MAE. Lower values are better. Bold and
underlined values denote the best and second-best results, respectively.}
\label{tab:spikf_full}
\scriptsize
\setlength{\tabcolsep}{2.2pt}
\renewcommand{\arraystretch}{1.04}
\resizebox{\textwidth}{!}{%
\begin{tabular}{@{}ll*{10}{cc}@{}}
\toprule
Dataset & Horizon
& \multicolumn{2}{c}{\textbf{SpikeLite}}
& \multicolumn{2}{c}{\textbf{SpikF}}
& \multicolumn{2}{c}{iTransformer}
& \multicolumn{2}{c}{RLinear}
& \multicolumn{2}{c}{PatchTST}
& \multicolumn{2}{c}{Crossformer}
& \multicolumn{2}{c}{TimesNet}
& \multicolumn{2}{c}{DLinear}
& \multicolumn{2}{c}{SCINet}
& \multicolumn{2}{c}{Autoformer}\\
\cmidrule(lr){3-4}\cmidrule(lr){5-6}\cmidrule(lr){7-8}
\cmidrule(lr){9-10}\cmidrule(lr){11-12}\cmidrule(lr){13-14}
\cmidrule(lr){15-16}\cmidrule(lr){17-18}\cmidrule(lr){19-20}
\cmidrule(l){21-22}
& & MSE & MAE & MSE & MAE & MSE & MAE & MSE & MAE
& MSE & MAE & MSE & MAE & MSE & MAE & MSE & MAE
& MSE & MAE & MSE & MAE \\
\midrule
ECL & 96  & .145 & .236 & .156 & .252 & .148 & .240 & .201 & .281 & .181 & .270 & .219 & .314 & .168 & .272 & .197 & .282 & .247 & .345 & .201 & .317 \\
    & 192 & .161 & .250 & .169 & .262 & .162 & .253 & .201 & .283 & .188 & .274 & .231 & .322 & .184 & .289 & .196 & .285 & .257 & .355 & .222 & .334 \\
    & 336 & .175 & .264 & .188 & .281 & .178 & .269 & .215 & .298 & .204 & .293 & .246 & .337 & .198 & .300 & .209 & .301 & .269 & .369 & .231 & .338 \\
    & 720 & .218 & .300 & .219 & .306 & .225 & .317 & .257 & .331 & .246 & .324 & .280 & .363 & .220 & .320 & .245 & .333 & .299 & .390 & .254 & .361 \\
\multicolumn{2}{l}{ECL Avg.} & \textbf{.175} & \textbf{.263} & .183 & .275 & \underline{.178} & \underline{.270} & .219 & .298 & .205 & .290 & .244 & .334 & .192 & .295 & .212 & .300 & .268 & .365 & .227 & .338 \\
\midrule
Weather & 96  & .158 & .196 & .163 & .200 & .174 & .214 & .192 & .232 & .177 & .218 & .158 & .230 & .172 & .220 & .196 & .255 & .221 & .306 & .266 & .336 \\
        & 192 & .209 & .243 & .209 & .241 & .221 & .254 & .240 & .271 & .225 & .259 & .206 & .277 & .219 & .261 & .237 & .296 & .261 & .340 & .307 & .367 \\
        & 336 & .266 & .284 & .266 & .283 & .278 & .296 & .292 & .307 & .278 & .297 & .272 & .335 & .280 & .306 & .283 & .335 & .309 & .378 & .359 & .395 \\
        & 720 & .346 & .337 & .344 & .334 & .358 & .347 & .364 & .353 & .354 & .348 & .398 & .418 & .365 & .359 & .345 & .381 & .377 & .427 & .419 & .428 \\
\multicolumn{2}{l}{Weather Avg.} & \textbf{.245} & \textbf{.265} & \textbf{.245} & \textbf{.265} & \underline{.258} & .278 & .272 & .291 & .259 & .281 & .259 & .315 & .259 & .287 & .265 & .317 & .292 & .363 & .338 & .382 \\
\midrule
ETTh1 & 96  & .383 & .391 & .379 & .391 & .386 & .405 & .386 & .395 & .414 & .419 & .423 & .448 & .384 & .402 & .386 & .400 & .654 & .599 & .449 & .459 \\
      & 192 & .437 & .422 & .432 & .421 & .441 & .436 & .437 & .424 & .460 & .445 & .471 & .474 & .436 & .429 & .437 & .432 & .719 & .631 & .500 & .482 \\
      & 336 & .472 & .445 & .473 & .441 & .487 & .458 & .479 & .446 & .501 & .466 & .570 & .546 & .491 & .469 & .481 & .459 & .778 & .659 & .521 & .496 \\
      & 720 & .478 & .460 & .474 & .459 & .503 & .491 & .481 & .470 & .500 & .488 & .653 & .621 & .521 & .500 & .519 & .516 & .836 & .699 & .514 & .512 \\
\multicolumn{2}{l}{ETTh1 Avg.} & \underline{.443} & \underline{.429} & \textbf{.440} & \textbf{.428} & .454 & .447 & .446 & .434 & .469 & .454 & .529 & .522 & .458 & .450 & .456 & .452 & .747 & .647 & .496 & .487 \\
\midrule
ETTh2 & 96  & .294 & .336 & .290 & .336 & .297 & .349 & .288 & .338 & .302 & .348 & .745 & .584 & .340 & .374 & .333 & .387 & .707 & .621 & .346 & .388 \\
      & 192 & .366 & .385 & .367 & .385 & .380 & .400 & .374 & .390 & .388 & .400 & .877 & .656 & .402 & .414 & .477 & .476 & .860 & .689 & .456 & .452 \\
      & 336 & .409 & .418 & .414 & .420 & .427 & .432 & .415 & .426 & .426 & .433 & 1.043 & .731 & .452 & .452 & .594 & .541 & 1.000 & .744 & .482 & .486 \\
      & 720 & .426 & .438 & .416 & .436 & .427 & .445 & .420 & .440 & .431 & .446 & 1.104 & .763 & .462 & .468 & .831 & .657 & 1.249 & .838 & .515 & .511 \\
\multicolumn{2}{l}{ETTh2 Avg.} & \underline{.374} & \textbf{.394} & \textbf{.372} & \textbf{.394} & .383 & .407 & .374 & .398 & .387 & .407 & .942 & .684 & .414 & .427 & .559 & .515 & .954 & .723 & .450 & .459 \\
\midrule
ETTm1 & 96  & .311 & .345 & .317 & .345 & .334 & .368 & .355 & .376 & .329 & .367 & .404 & .426 & .338 & .375 & .345 & .372 & .418 & .438 & .505 & .475 \\
      & 192 & .369 & .377 & .372 & .372 & .379 & .391 & .391 & .392 & .367 & .385 & .450 & .451 & .374 & .387 & .380 & .389 & .439 & .450 & .553 & .496 \\
      & 336 & .396 & .395 & .401 & .394 & .426 & .420 & .424 & .415 & .399 & .410 & .532 & .515 & .410 & .411 & .413 & .413 & .490 & .485 & .621 & .537 \\
      & 720 & .473 & .439 & .461 & .430 & .491 & .459 & .487 & .450 & .454 & .439 & .666 & .589 & .478 & .450 & .474 & .453 & .595 & .550 & .671 & .561 \\
\multicolumn{2}{l}{ETTm1 Avg.} & \textbf{.387} & \underline{.389} & \underline{.388} & \textbf{.385} & .407 & .410 & .414 & .407 & .387 & .400 & .513 & .496 & .400 & .406 & .403 & .407 & .485 & .481 & .588 & .517 \\
\midrule
ETTm2 & 96  & .175 & .254 & .175 & .251 & .180 & .264 & .182 & .265 & .175 & .259 & .287 & .366 & .187 & .267 & .193 & .292 & .286 & .274 & .255 & .339 \\
      & 192 & .240 & .295 & .242 & .296 & .250 & .309 & .246 & .304 & .241 & .302 & .414 & .492 & .249 & .309 & .284 & .362 & .399 & .445 & .281 & .340 \\
      & 336 & .302 & .336 & .302 & .336 & .311 & .348 & .307 & .342 & .305 & .343 & .597 & .542 & .321 & .351 & .369 & .427 & .637 & .591 & .339 & .372 \\
      & 720 & .395 & .390 & .405 & .397 & .412 & .407 & .407 & .398 & .402 & .400 & 1.730 & 1.042 & .408 & .403 & .554 & .522 & .960 & .735 & .433 & .432 \\
\multicolumn{2}{l}{ETTm2 Avg.} & \textbf{.278} & \textbf{.319} & \underline{.281} & \underline{.320} & .288 & .327 & .286 & .327 & .281 & .326 & .757 & .610 & .291 & .333 & .350 & .401 & .571 & .537 & .327 & .371 \\
\midrule
Traffic & 96  & .453 & .285 & .477 & .286 & .395 & .268 & .649 & .389 & .462 & .295 & .522 & .290 & .593 & .321 & .650 & .396 & .788 & .499 & .613 & .388 \\
        & 192 & .468 & .287 & .481 & .289 & .417 & .276 & .601 & .366 & .466 & .296 & .530 & .293 & .617 & .336 & .598 & .370 & .789 & .505 & .616 & .382 \\
        & 336 & .481 & .293 & .499 & .295 & .433 & .283 & .609 & .369 & .482 & .304 & .558 & .305 & .629 & .336 & .605 & .373 & .797 & .508 & .622 & .337 \\
        & 720 & .514 & .311 & .533 & .312 & .467 & .302 & .647 & .387 & .514 & .322 & .589 & .328 & .640 & .350 & .645 & .394 & .841 & .523 & .660 & .408 \\
\multicolumn{2}{l}{Traffic Avg.} & \underline{.479} & \underline{.294} & .497 & .296 & \textbf{.428} & \textbf{.282} & .626 & .378 & .481 & .304 & .550 & .304 & .620 & .336 & .625 & .383 & .804 & .509 & .628 & .379 \\
\midrule
Exchange & 96  & .084 & .204 & .084 & .201 & .086 & .206 & .093 & .217 & .088 & .205 & .256 & .367 & .107 & .234 & .088 & .218 & .267 & .396 & .197 & .323 \\
         & 192 & .180 & .300 & .180 & .300 & .177 & .299 & .184 & .307 & .176 & .299 & .470 & .509 & .226 & .344 & .176 & .315 & .351 & .459 & .300 & .369 \\
         & 336 & .336 & .417 & .334 & .417 & .331 & .417 & .351 & .432 & .301 & .397 & 1.268 & .883 & .367 & .448 & .313 & .427 & 1.324 & .853 & .509 & .524 \\
         & 720 & .864 & .695 & .841 & .690 & .847 & .691 & .886 & .714 & .901 & .714 & 1.767 & 1.068 & .964 & .746 & .839 & .695 & 1.058 & .797 & 1.447 & .941 \\
\multicolumn{2}{l}{Exchange Avg.} & .366 & .404 & \textbf{.360} & \textbf{.402} & \underline{.360} & \textbf{.403} & .378 & .417 & .367 & \underline{.404} & .940 & .707 & .416 & .443 & \textbf{.354} & .414 & .750 & .626 & .613 & .539 \\
\bottomrule
\end{tabular}}
\end{table*}